\documentclass[sigchi,nonacm]{acmart}
 
\RequirePackage{snapshot}

\renewcommand\footnotetextcopyrightpermission[1]{} 
\usepackage[utf8]{inputenc} 
\usepackage[T1]{fontenc}    
\usepackage{url}            
\usepackage{amsfonts}       
\usepackage{microtype}      

\usepackage{graphicx}
\usepackage{color}
\usepackage{hyperref}
\usepackage[all]{hypcap}  

\usepackage{subcaption}
\usepackage{cuted}
\usepackage{booktabs}       
\usepackage{url}

\renewcommand{\eqref}[1]{(\ref{eq:#1})}
\newcommand{\secref}[1]{\S\ref{sec:#1}}
\newcommand{\figref}[1]{Fig.~\ref{fig:#1}}

\newcommand{\aphours}{420}

\newcommand{\apcars}{16}

\newcommand{\website}{\url{https://hcai.mit.edu/arguing-machines}}

\usepackage{fancyhdr}
\fancypagestyle{firststyle}
{
  \fancyhf{}
  \fancyfoot[L]{
    \vspace{0.1in}
    * Corresponding author. Email: \texttt{fridman@mit.edu}.
  }
}

\begin{document}

\title{Arguing Machines: Human Supervision of Black Box AI Systems That Make Life-Critical Decisions}

\newcommand{\authspace}{\hspace{0.15in}}
\author{Lex Fridman* \authspace Li Ding \authspace Benedikt Jenik \authspace Bryan Reimer}
\affiliation{%
  \institution{Massachusetts Institute of Technology (MIT)}
}

\renewcommand{\shortauthors}{L. Fridamn et al.}
\renewcommand{\shorttitle}{Arguing Machines: Human Supervision of Black Box AI Systems}

\begin{abstract}
We consider the paradigm of a black box AI system that makes life-critical decisions. We propose an ``arguing machines'' framework that pairs the primary AI system with a secondary one that is independently trained to perform the same task. We show that disagreement between the two systems, without any knowledge of underlying system design or operation, is sufficient to arbitrarily improve the accuracy of the overall decision pipeline given human supervision over disagreements. We demonstrate this system in two applications: (1) an illustrative example of image classification and (2) on large-scale real-world semi-autonomous driving data.  For the first application, we apply this framework to image classification achieving a reduction from 8.0\% to 2.8\% top-5 error on ImageNet. For the second application, we apply this framework to Tesla Autopilot and demonstrate the ability to predict 90.4\% of system disengagements that were labeled by human annotators as challenging and needing human supervision.
\end{abstract}

\begin{teaserfigure}
  \centering
  \includegraphics[width=\textwidth]{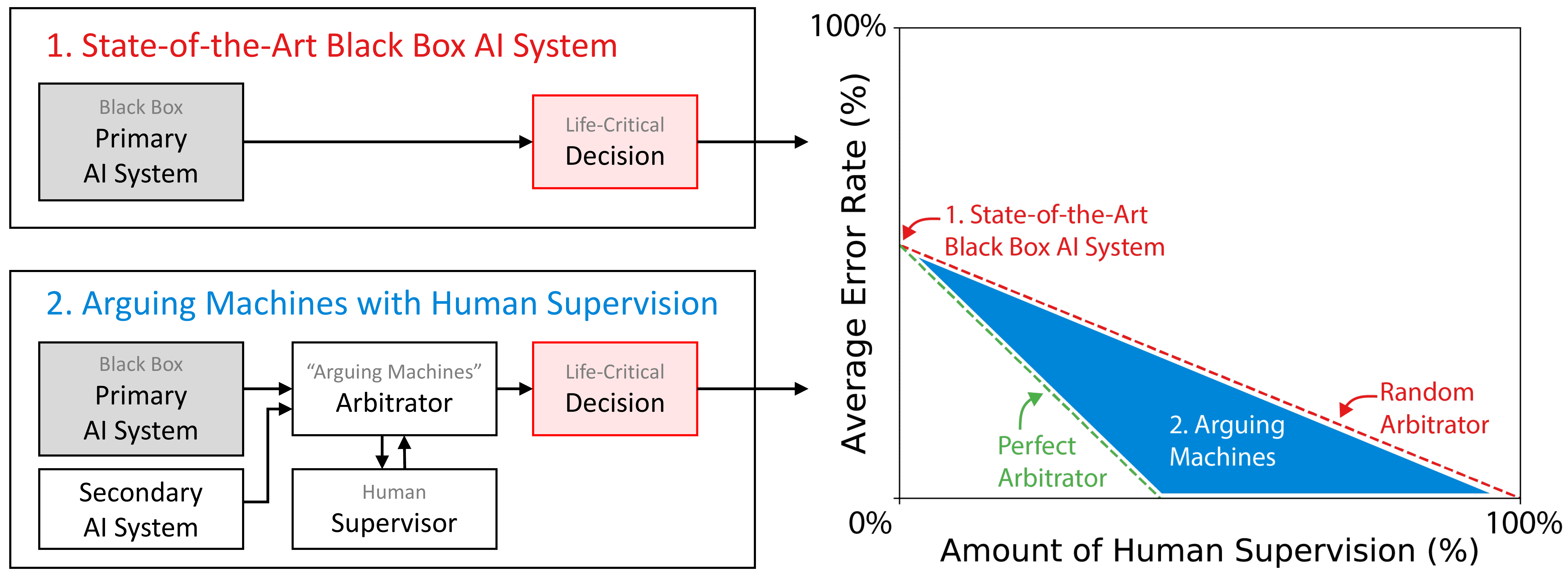}
  \captionof{figure}{``Arguing machines'' framework that adds a secondary system to a primary ``black box'' AI system
    that makes life-critical decisions and uses disagreement between the two as a signal to seek human supervision. We
    demonstrate that this can be a powerful way to reduce overall system error.}
  \label{fig:concept}
\end{teaserfigure}

\maketitle
\thispagestyle{firststyle} 

\begin{figure*}[ht!]
  \centering 
  \includegraphics[width=\textwidth]{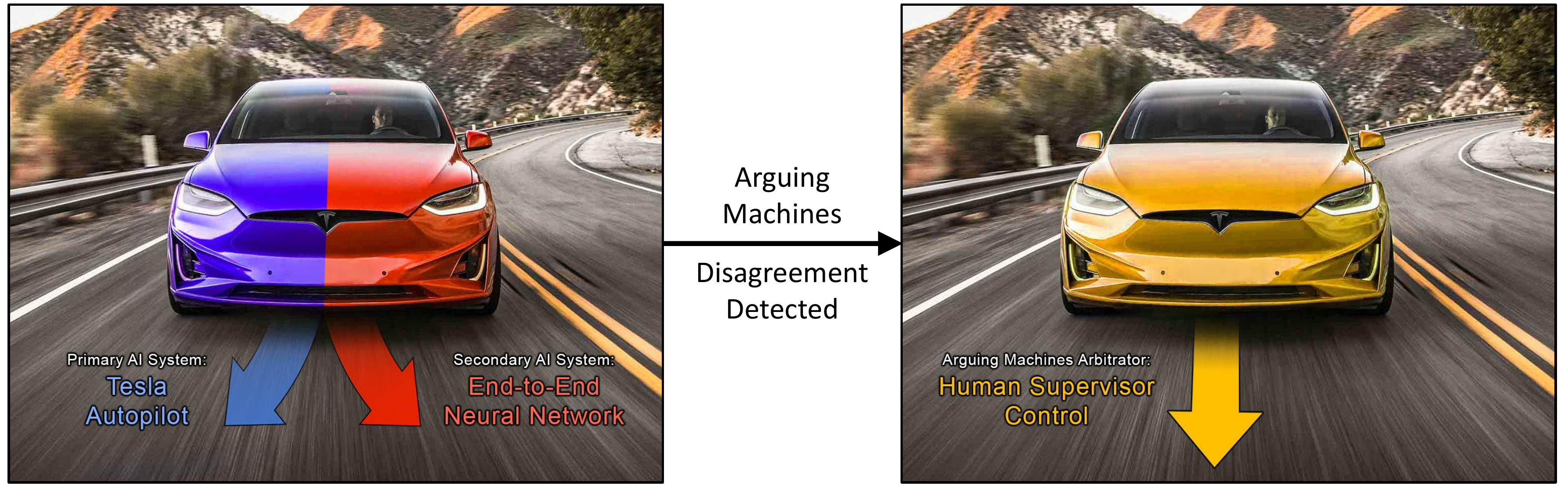}
  \caption{Concept diagram of the arguing machines framework applied to the automated steering task where the primary
    system is Tesla Autopilot and the secondary system is an end-to-end neural network. When disagreement between the
    two exceed a threshold, the human supervisor is notified and may elect to take control of the vehicle.}
  \label{fig:split-car}
\end{figure*}

\section{Introduction}\label{sec:introduction}

Successful operation of intelligent automated systems in real-world applications where errors are assigned extremely
high costs, such as when the systems are tasked with making life-critical decisions, is one of the grand challenges
facing the AI community. The difficulty is not within the task itself, but rather in the small margin of allowable error
given the human life at stake and the large number of edge cases that have to be accounted for in real-world
operation. This challenge has two categories of approaches: (1) improve the accuracy of the system such that it reaches
the acceptable level of performance, or (2) integrate the system with a human supervisor that aids its operation such
that the combined system of human and machine reach the acceptable level of performance. The former set of approaches
has been the focus of the machine learning community. The latter is the focus of this paper.

We consider the real-world operating paradigm of a black box AI system (termed ``primary system'') that is tasked with
making life-critical decisions. The proposed method integrates the human being into the critical role of resolving
uncertainty and disagreement in decisions whose errors are associated with high negative utility values.
We demonstrate this system in two applications: (1) an illustrative example of image classification and
(2) on large-scale real-world semi-autonomous driving data.  For the first application, we show this framework applied
to image classification achieving an improvement from 8.0\% to 2.8\% top-5 error on ImageNet over ResNet-50 network
(treated as a black box).  For the second application, we apply the arguing machines framework to monocular-vision-based
automated steering systems. The first is a proprietary Tesla Autopilot system equipped in the first generation of
Autopilot-capable vehicles. The second is an end-to-end neural network trained on a large-scale naturalistic dataset of
420 hours or 45 million frames of autonomous driving in Tesla vehicles. We demonstrate the ability of the overall
arguing machines to predict 90.4\% of system disengagements that were deemed as ``tricky'' by human annotators and thus
likely to be associated with high-probability of driver injury if not handled by the driver.

This paper demonstrates the surprising and impactful finding that the disagreement between two systems, without any
knowledge of the design of either system, may have sufficient information to significantly improve the performance of
the overall framework when combined with human supervision. This result has serious implications for the design of
effective and safe human-computer interaction experiences.

\subsection{Arguing Machines Concept}\label{sec:concept}

The ``black box'' nature of AI systems is the property of some machine learning approaches that make it difficult to
``see inside'' the model inference process that makes a particular decision. This is both due to the inherent difficult
of engineering \textit{explainable AI} systems \cite{gunning2017explainable} and the natural reluctance by companies
that provide the AI system to visualize the inner workings of the system and to reveal uncertainty of predictions and
system errors. The motivation for this work is that there are applications in which such errors can lead to loss of
human life. Errors are inherently part of supervised machine learning systems that seek to generalize from patterns of
the past to pattern of the present. It is very difficult to engineer such errors out completely. We propose to instead
manage them by integrating the human being as a supervisor. This is important for both creating a safe
interaction with an AI system, but also a more effective human-computer interaction experience that develops an
appropriate amount of trust and understanding.

\figref{concept} shows the \textit{arguing machines} framework. Consider that there is a primary AI system trained to
perform a specific task. A task is defined as making a decision based on a well-defined input. For image classification
(see \secref{imagenet}), the task is to take an image as input and make a prediction of likelihood that the image is one
of a number of categories. For autonomous steering (see \secref{driving}), the task is to take a sequence of video
frames of the forward roadway and make a steering decision. The output of this system is a decision, discrete in the
former case and continuous in the latter case. The arguing machines framework introduces a secondary system trained to
perform the same task without any interaction with the primary system. The disagreement between the two systems is
measured by the arbitrator and passed to a human supervisor if the disagreement exceeds a constant predefined
threshold. This threshold controls the tradeoff between the relative amount of human supervision and overall system
error as illustrated in \figref{concept}.






\subsection{Real-World Application: Autonomous Driving}

We use image classification in \secref{imagenet} as an illustrative case study to demonstrate the concept of arguing
machines. However, in this work, the central case study of applying the \textit{arguing machines} framework in the real
world is semi-autonomous driving (detailed in \secref{driving}). We chose this application because it is a domain where
AI systems are already making hundreds of thousands of life-critical decisions every day in Tesla vehicles equipped with
Autopilot \cite{fridman2018apmiles} and many other cars equipped with various degrees of automation
\cite{fridman2018avt}. These perception-control systems are black box AI systems that provide very limited communication
of system limits, uncertainty, and errors to the driver. Therefore, we believe applying the arguing machines framework
in this context may help integrate the human driver in a way that may help save their life.

As shown in \figref{split-car}, for the semi-autonomous driving case study, the role of the primary machine is served by
the first generation of Tesla Autopilot software with the perception and steering predictions performed by the
integrated Mobileye system \cite{pirzada2015autopilot}. The role of the secondary machine in this paper is served by an
end-to-end convolutional neural network similar to that described and evaluated in \cite{bojarski2016end} except that
our model considers the temporal dynamics of the driving scene by taking as input some aspects of the visual change in
the forward-facing video for up to 1 second back in time (see \secref{network}). The output of both systems is a
steering angle. The differences in those outputs is what constitutes the argument based on which disengagement
suggestions and edge case proposals are made. The network model is trained on a balanced dataset constructed through
sampling from \aphours~hours of real-world on-road automated driving by a fleet of \apcars~Tesla vehicles
\cite{fridman2018avt} (see \secref{dataset}).

The central idea proposed in this work is that robustness of the artificial intelligence system behind the perception
and planning necessary for automated driving can be achieved by supplementing the training dataset with edge cases
automatically discovered through monitoring the disagreement between multiple machine learning models.

\begin{figure*}[!ht]
  \centering
  \includegraphics[width=\textwidth]{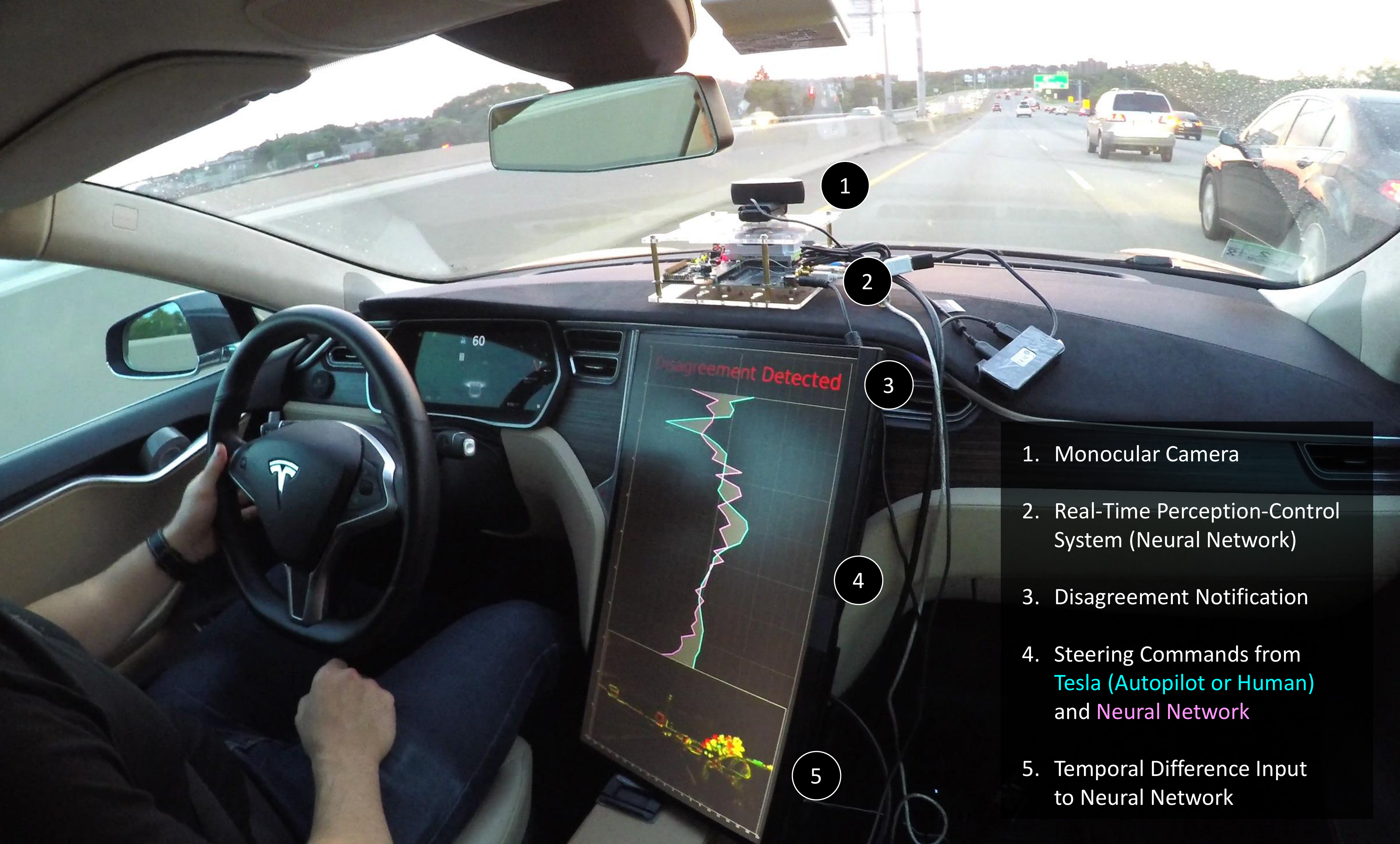}
  \caption{Implementation and evaluation of the system presented in this paper. The primary perception-control system is
    Tesla Autopilot. The secondary perception-control system is an end-to-end neural network. We equipped a Tesla Model
    S vehicle with a monocular camera, an NVIDIA Jetson TX2, and an LCD display that shows the steering commands from
    both systems, the temporal difference input to the neural network, and (in red text) a notice to the driver when a
    disagreement is detected.}
  \label{fig:demo}
\end{figure*}

We implement and deploy the system described in this work to show its capabilities and performance in real-world
conditions. Its successful operation is exhibited in an extensive, on-road video demonstration that is made publicly
available at \website. As \figref{demo} shows, we instrumented a Tesla Model S vehicle with an NVIDIA Jetson
TX2 running the neural network based perception-control system and disagreement function in real-time. The input to the
system is a forward-facing monocular camera and the output are steering commands. The large display shows steering
commands both from the primary system (Tesla) and secondary system (neural network), and notifies the driver when a
disagreement is detected.

The case studies presented in this paper have associated data, source code, and demonstration videos that are made available on
\website.

\section{Related Work}

Life-critical and safety-critical systems are those whose failure may result in loss of human life
\cite{knight2002safety}. Naturally, many domains of real-world human-machine interaction involve risk of injury and loss
of life through a long sequence of cause and effect that is far removed from the initial decisions made by the
machine. In this work, we are focusing on applications where a single erroneous decision by an AI system has a
high-likelihood of causing direct harm to a human being in a way that does not separate the initial decision from the
final negative result via a chaos of unintended consequences. This latter paradigm is less amenable to analysis
\cite{hilborn2004sea}.

The real-world application data analyzed in this work is from the domain of autonomous vehicle perception-control
systems. Other application domains where AI systems make life-critical decision include medicine, nuclear engineering,
aviation, and autonomous weapon systems. Medical diagnosis is the process in medicine that is clearly amenable to
assistance by AI systems, assuming the specific diagnosis task can be formalized and digitally grounded in human
measurement data. In many cases, this process is life-critical in that a misdiagnosis (incorrect diagnosis) can lead to
bodily harm and loss of life \cite{kendrey1996misdiagnosis}. Such a diagnosis task can be directly formed into an exam
classification problem, allowing for supervised deep learning methods to be effectively applied. In exam classification,
one or multiple images (an exam sample) as input is matched with a single diagnostic variable as output (e.g., disease
present or not).  \cite{gulshan2016development} applies deep learning to create an algorithm for automated detection of
diabetic retinopathy and diabetic macular edema in retinal fundus photographs.  \cite{esteva2017dermatologist}
demonstrates classification of skin lesions using a single CNN, trained end-to-end from images directly to predict
disease labels.

\subsection{Ensemble of Neural Networks}

The idea of multiple networks collaborating or competing against each other to optimize an objective have been
implemented in various contexts. For example, multiple networks have been combined together in order to improve accuracy
\cite{krogh1995neural} as have traditionally been explored in machine learning as ensembles of classifiers. 
For deep neural networks, \cite{srivastava2014dropout} propose a technique that provides a way of approximately 
combining exponentially many different network architectures.
Recent work \cite{he2016deep} combine six models of different depth to form an ensemble. 
\cite{szegedy2015going}  independently trained seven versions of the same network with same initialization, 
which only differ in sampling methodologies and the randomized input image order. 
In these approaches, decision-level fusion is performed across many classifiers in order to increase accuracy and 
robustness of the overall system. 

Besides, ensemble can also be done on the dataset-level. 
Early statistical sampling methods such as \cite{efron1992bootstrap} can be used to improve the performance and 
get the confidence interval of a model. 
\cite{everingham2015pascal, russakovsky2015imagenet} use the method to test whether the performance of different 
networks is statistically significantly different, and obtain the confidence interval of error rate. 
Moreover for computer vision specifically, various ensemble methods can be done on input-level, such as averaging 
prediction of five different crops and their horizontal reflections~\cite{krizhevsky2012imagenet}, multi-scale multi-crop 
prediction~\cite{szegedy2015going, he2016deep, simonyan2014very}, are commonly used to increase accuracy and 
robustness of the whole system during testing. 
However, \cite{szegedy2015going} also note that such terminology may not be necessary in real-world applications, 
as the benefit of which becomes marginal after a reasonable number.

Alternatively, generative adversarial networks (GANs)~\cite{goodfellow2014generative} have two different networks working 
against each other for representation learning and subsequent generation of samples from those learned representations, 
including generation of steering commands \cite{kuefler2017imitating}. Neural networks have also been used in different
environments at the same time \cite{mnih2016asynchronous} to learn from them in parallel, or, as in our work, to look at
what the disagreement to other systems reveals about the underlying state of the world the networks operate in. 
Although not directly referred in our work, the above research share the similar idea of using the disagreement between 
different systems, and indicates that there is much information contained in such disagreement.

\subsection{End-to-End Approaches to Driving}

In contrast to modular engineering approaches to self-driving systems, where deep learning only plays a role for the
initial scene interpretation step~\cite{huval2015empirical}, it is also possible to approach driving as a more holistic task that can possibly be
solved in a data-driven way by a single learner: an end-to-end neural network. First attempts were made almost 30 years
ago \cite{pomerleau1989alvinn}, long before the recent GPU-enabled performance breakthroughs in deep learning \cite{krizhevsky2012imagenet}.

A similar, but more modern approach using deeper, convolutional nets has been deployed in an experimental vehicle by
NVIDIA \cite{bojarski2016end}, and further improvements to that were made using various forms of data augmentation
\cite{ross2011reduction} and adapted to the driving context by \cite{zhang2017query}.  A more advanced approach
\cite{xu2016end} formulates autonomous driving as a vehicle egomotion prediction problem, and uses an end-to-end sequence
model built upon a scene perception model. They also show that by training scene perception alone as a side task further
improves the whole system.  Recently, \cite{kim2017interpretable} studies the visual explanations and network's behavior
in end-to-end driving, by using a visual attention model to train a convolutional network from images to steering angle.

\newcommand{\imagenetfig}[1]{%
  \begin{subfigure}[t]{0.9\columnwidth}
    \setlength{\fboxsep}{0pt}%
    \setlength{\fboxrule}{0.3pt}%
    \fbox{\includegraphics[width=\textwidth]{images/imagenet/examples7_final/imagenet#1.jpg}}%
    \label{fig:imagenet-#1}
  \end{subfigure}\hspace{0.06in}
}
\newcommand{\imagenetspace}{\vspace{0.1in}}
\begin{figure*}[tbp]
  \centering
  \imagenetfig{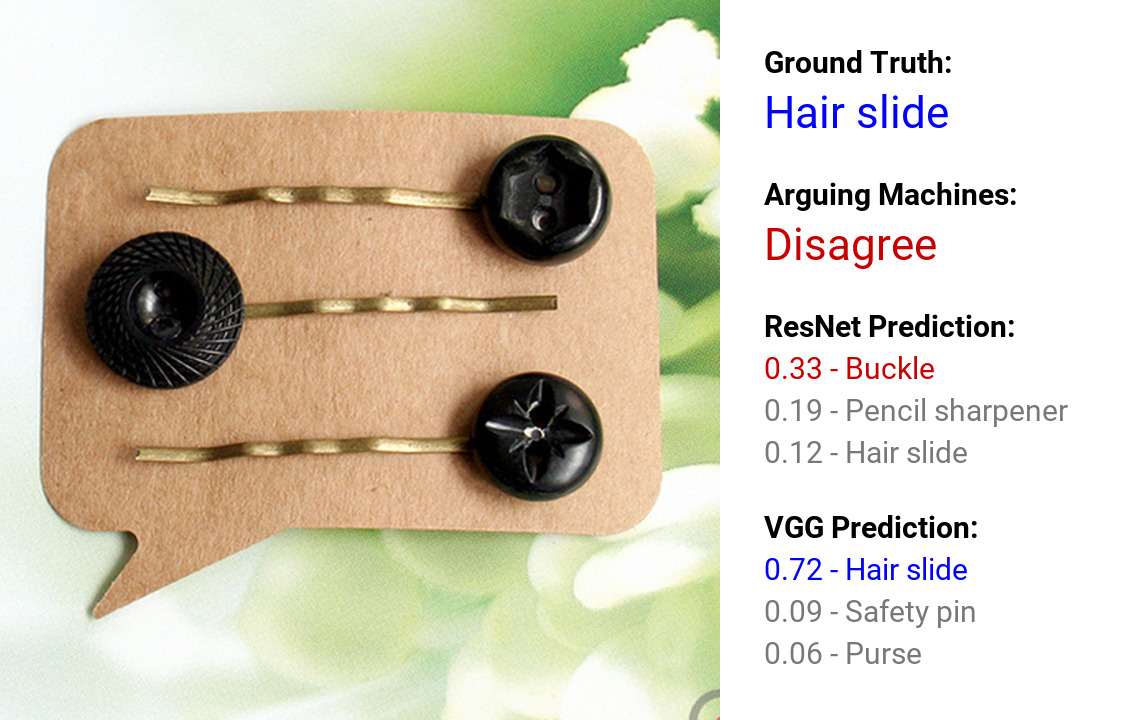}
  \imagenetfig{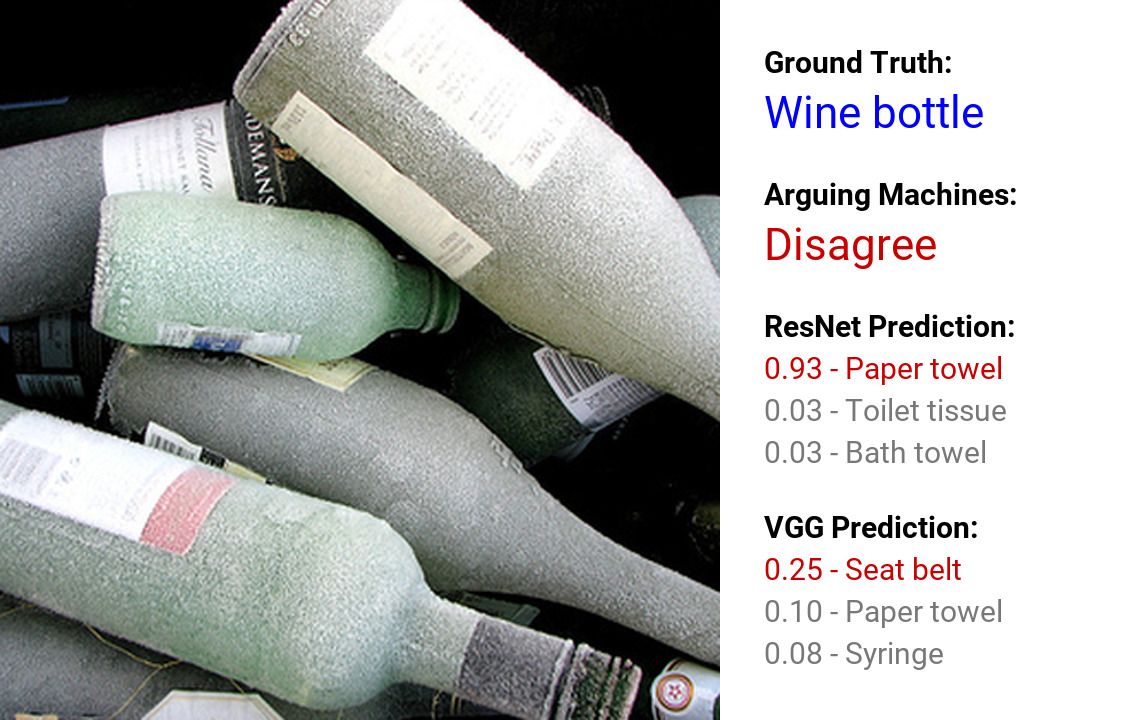}\\\imagenetspace
  \imagenetfig{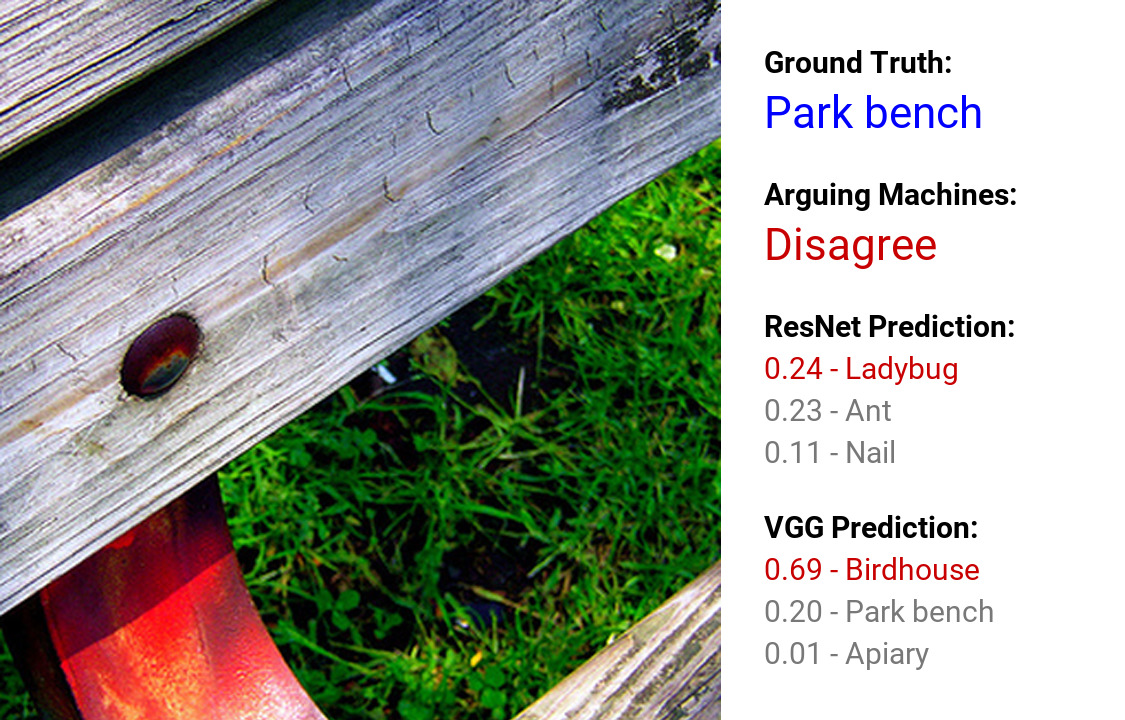}
  \imagenetfig{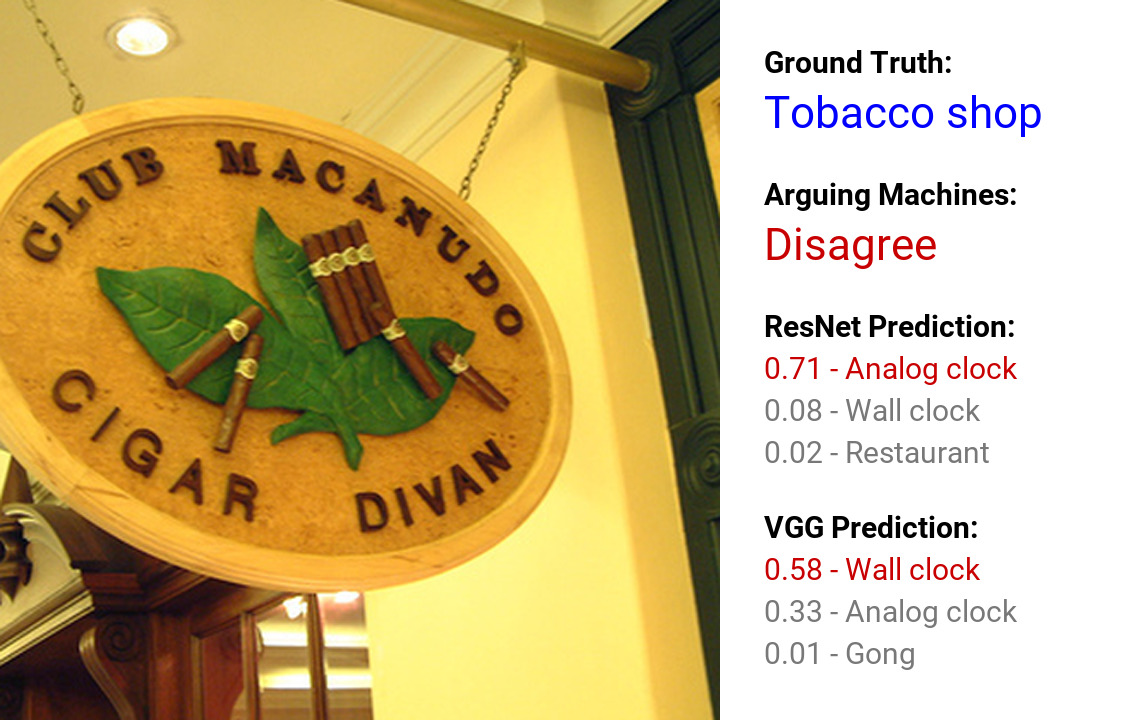}\\\imagenetspace
  \imagenetfig{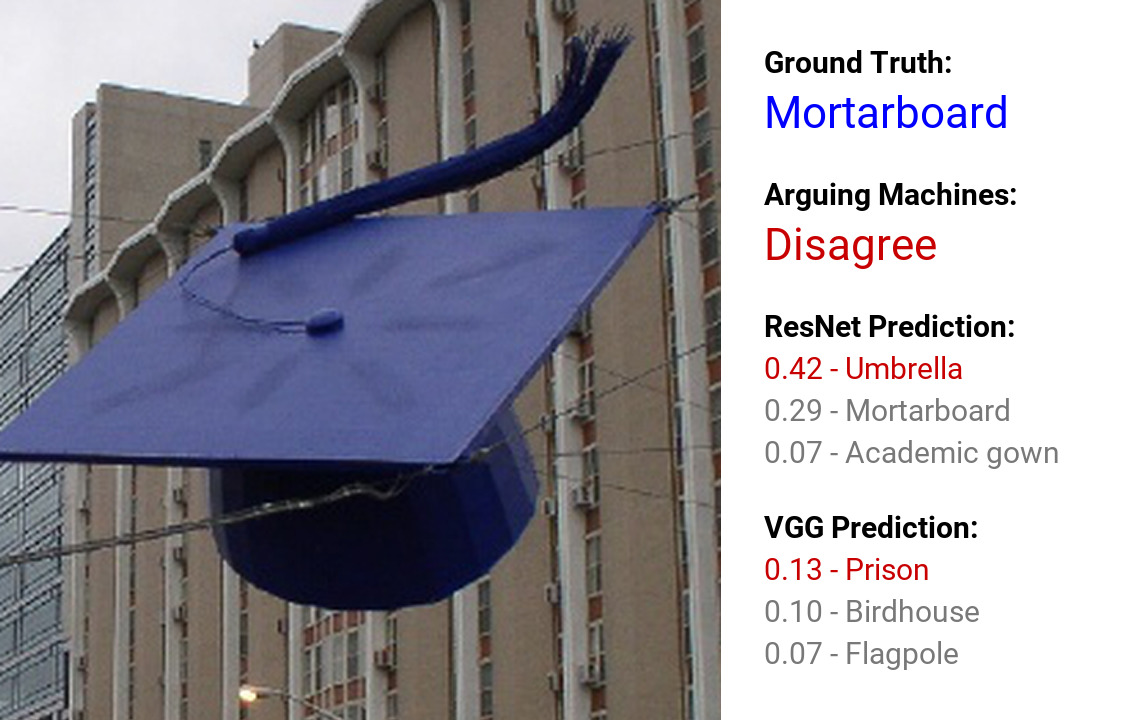}
  \imagenetfig{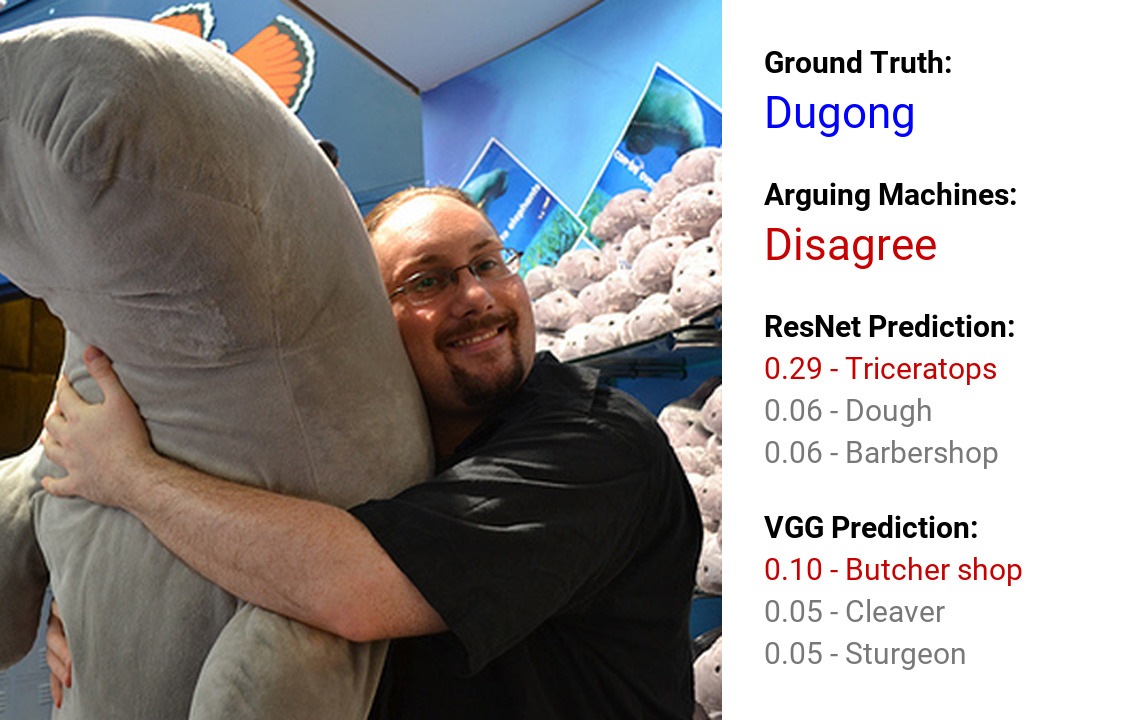}\\\imagenetspace
  \imagenetfig{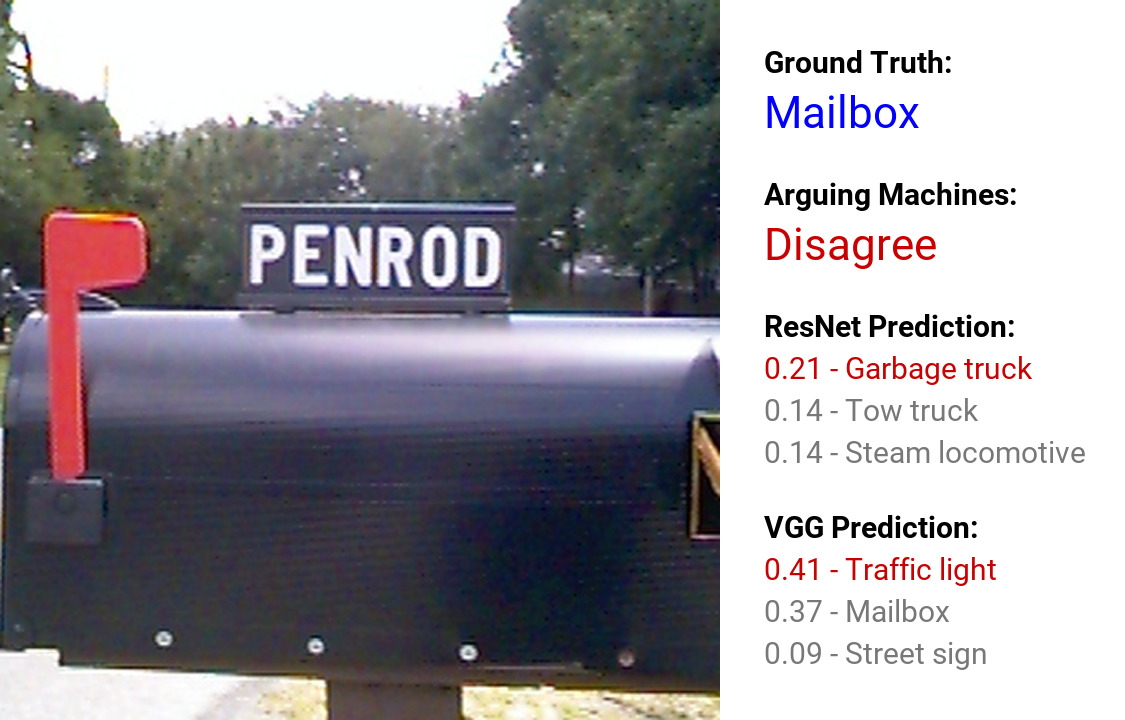}
  \imagenetfig{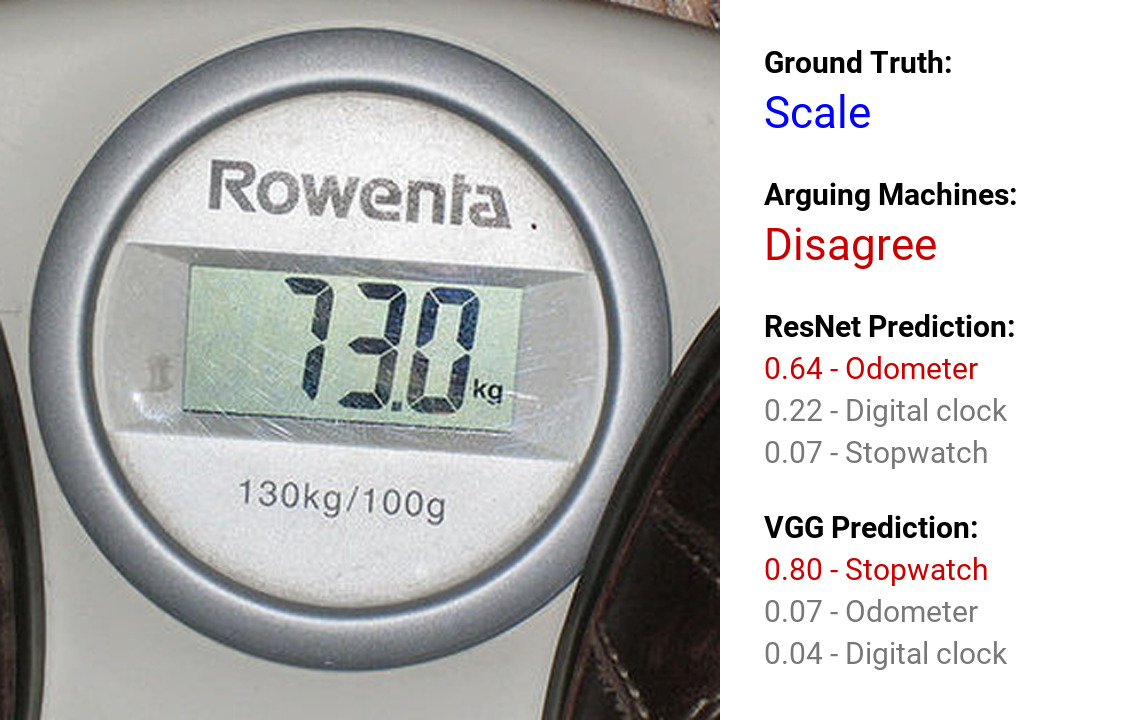}\\
  \caption{ImageNet examples where the primary system (ResNet) and secondary system (VGG) disagree on the image
    classification task. The ground truth and correct classifications are shown in blue. Incorrect classifications are
    shown in red. }
  \label{fig:imagenet}
\end{figure*}

\section{Arguing Machines for Image Classification on ImageNet}\label{sec:imagenet}

The ImageNet Dataset~\cite{deng2009imagenet} and Challenge~\cite{russakovsky2015imagenet} has become the standard
benchmark for large-scale object recognition, allowing significant algorithmic advances in large-scale image recognition
and retrieval. Most of the state-of-the-art approaches~\cite{krizhevsky2012imagenet, simonyan2014very, he2016deep} are
variants of deep convolutional neural network architectures. However, although significant strides toward solving the
image classification problem have been taken, the systems are still far from perfection. We chose image classification
as the illustrative case study because it is one of the best studied problems in artificial intelligence, and yet even
in this well-studied problem space, we can demonstrate improvement by integrating human supervision via the arguing
machines framework.

If we consider the general process of decision making, aggregating ideas from multiple sources strengthens the
generalizability of the decision. A single source is likely to be biased due to factors of data selection or underlying
model specifics. This concept is widely used in machine learning algorithms to improve performance. Despite the fact
that deep neural networks models themselves are ensembles of linear functions with non-linear activations, unsupervised
ensemble methods such as bootstrap~\cite{efron1992bootstrap}, bagging~\cite{breiman1996bagging},
dropout~\cite{srivastava2014dropout} and supervised ones such as stacking~\cite{wolpert1992stacked, breiman1996stacked}
can be utilized to improve the generalization accuracy of the overall system.

In this paper we consider the idea that in collaborative decision making, disagreement may contain as much if not more
critical information than agreement, especially when the individual decision makers are very good at the task in
question.  We explore this kind of disagreement in a machine learning scenario, and seek to leverage the information behind
such disagreement in order to improve the overall performance of the system..

In this section, we illustrate the idea of arguing machines with a toy experiment on ImageNet Dataset. The arguing
machines framework is proposed as follows. Suppose, there exists a state-of-the-art black-box AI system (primary system)
whose accuracy is great but not perfect. In order to safely use or test the system, we propose to have a secondary system
that can argue with the primary system. When disagreement arises between two systems, we regard it as a difficult case
and mark it as needing human supervision. The purpose of arguing machines is to improve the system performance with minimal
human effort, especially when the primary system is a black-box and gives no other information except the final output.

The experiment in this section is a common image classification task. We take two popular image recognition models,
VGG~\cite{simonyan2014very} and ResNet~\cite{he2016deep}. Specifically, we treat a single ResNet-50 model as the
black-box and a VGG-16 model as an end-to-end deep learning model. The models are pre-trained and we obtain the
prediction results from single center-cropped images in the ImageNet validation set.

The arguing machines arbitrator detects the disagreement when the top predictions of two systems differ.  In this experiment, ResNet
and VGG disagree on $11645$ images, which is $23.3\%$ of the whole validation set.  For the results of arguing machines,
we assume with human taking look at the disagreement cases, the classification is always correct.  We also propose a
baseline method that with the same amount of images send to human verification (always correct), but randomly selected.
We evaluate both the top-1 error and the top-5 error. The results are shown in Table~\ref{tb:imgnet}.

\newcommand{\topone}{\begin{tabular}[c]{@{}l@{}}Top-1\\ Error (\%)\end{tabular}}
\newcommand{\topfive}{\begin{tabular}[c]{@{}l@{}}Top-5\\ Error (\%)\end{tabular}}
\begin{table}[!ht]
  \caption{Experimental results on ImageNet-val set.}
  \label{tb:imgnet}
  \centering
  \begin{tabular}{lll}
    \toprule
    Method     &  \topone  &  \topfive  \\
    \midrule
    ResNet-50 (primary system) & 25.2 & 8.0 \\
    VGG-16 (secondary system) & 29.0 & 10.1 \\
    Ensemble: ResNet-50, VGG-16 & 24.4 & 7.8 \\
    \midrule
    Random Arbitrator & 19.3 & 6.2 \\
    Arguing Machines & \textbf{10.7} & \textbf{2.8} \\
    \bottomrule
  \end{tabular}
\end{table}

The results show that with the arguing machines framework, the performance of a state-of-the-art image
recognition system can be significantly improved, even when we treat it as a black-box system.

Table~\ref{tb:imgnetdis} shows the analysis of arguing machines in this context.  With less than a quarter of images
verified by a human supervisor, the arguing machines framework is able to detect more than half of the failure cases in
both top-1 and top-5 tasks, even given the fact that both systems already have very strong performance.  Such results
also indicate that although two deep convolutional neural networks are trained on the same dataset, with similar
architectures featuring a combination of convolutional layers, fully connected layers, dropout layers, etc., the behavior of
the two trained systems is quite different, as they do not fail the same way during testing. This is a surprising and
fascinating result that reveals the predictive power of disagreement between artificial intelligence systems.

The precision of top-5 classification is much lower than top-1, because the two systems can be both correct even if they
disagree on the top prediction. However the recall for both top-1 and top-5 tasks are consistently high, indicating that
even with the simpler classification task, where systems fail less often, the arguing machines framework can still detect many of the failure
cases with disagreements and in so doing significantly reduce the error.

\begin{table}
  \caption{Performance analysis of arguing machines.}
  \label{tb:imgnetdis}
  \centering
  \begin{tabular}{lll}
    \toprule
    Task     &  Precision (\%) &  Recall (\%)  \\
    \midrule
    Top-1 Classification & 62.4 & 57.6 \\
    Top-5 Classification & 22.2 & 64.6 \\
    \bottomrule
  \end{tabular}
\end{table}

Examples of disagreements between the primary and secondary systems on the image classification task are shown in
\figref{imagenet}. More examples, including disagreement over object detection and classification in video, are available online at \website.

\section{Arguing Machines for Semi-Autonomous Driving}\label{sec:driving}

Software is taking on greater operational control in modern vehicles and in so doing is opening the door to machine
learning. These approaches are fundamentally hungry for data, based on which, they aim to take on the higher level
perception and planning tasks. As an example, over 15 million vehicles worldwide are equipped with Mobileye computer
vision technology \cite{stein2015collision}, including the first generation Autopilot system that serves as the
``primary machine'' in this work.

Given the requirement of extremely low error rates and need to generalize over countless edge cases, large-scale
annotated data is essential to making these approaches work in real-world conditions. In fact, for driving, training
data representative of all driving situations may be more important than incremental improvements in perception,
control, and planning algorithms. Tesla, as an example, is acknowledging this need by asking its owners to
share data with the company for the explicit purpose of training the underlying machine learning models. Our work does
precisely this, applying end-to-end neural network approaches to training on large-scale, semi-autonomous, real-world
driving data. The resulting model serves as an observer and critic of the primary system with the goals of (1) discovering
edge cases in the offline context and (2) bringing the human back into the loop when needed in the online context.

\begin{figure}[h!]
  \centering
  \includegraphics[width=\columnwidth]{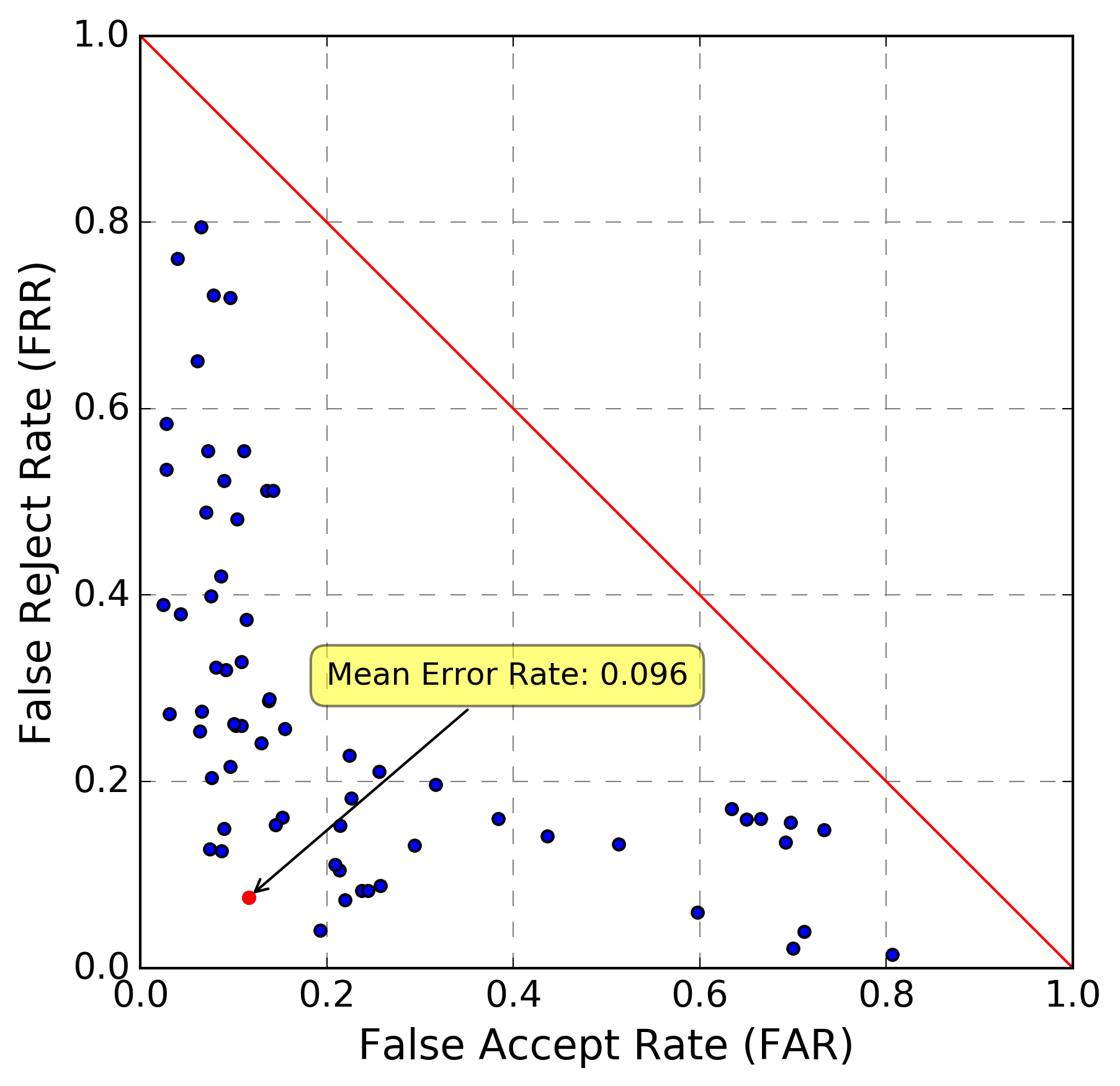}
  \caption{The tradeoff between false accept rate (FAR) and false reject rate (FRR) achieved by varying the constant
    threshold used to make the binary disagreement classification. The red circle designates a threshold of 10 that is
    visualization on an illustrative example in \figref{disagreement-example}.}
  \label{fig:toc-roc}
\end{figure}

We perform two evaluations in our application of arguing machines to semi-autonomous driving. First, we evaluate the
ability of the end-to-end network to predict steering angles commensurate with real-world steering angles that were used
to keep the car in its lane. For this, we use distinct periods of automated lane-keeping during Autopilot engagement as
the training and evaluation datasets. Second, we evaluate the ability of an argument arbitrator (termed ``disagreement
function'') to estimate, based on a short time window, the likelihood that a transfer of control is initiated, whether
by the human driver (termed ``human-initiated'') or the Autopilot system itself (termed ``machine-initiated''). We have
6,500 total disengagements in our dataset. All disengagements (whether human-initiated or machine-initiated) are
considered to be representative of cases where the visual characteristics of the scene (e.g., poor lane markings,
complex lane mergers, light variations) were better handled by a human operator. Therefore, we chose to evaluate the
disagreement function by its ability to predict these disengagements, which it is able to do with 90.4\% accuracy (see
\figref{toc-roc}).

\subsection{Naturalistic Driving Dataset}\label{sec:dataset}

The dataset used for the training and evaluation of the end-to-end steering network model comprising the ``secondary
machine'' is taken from a large-scale naturalistic driving study of semi-autonomous vehicle technology
\cite{fridman2018avt}. Specifically, we used 420 hours of driving data where a Tesla Autopilot system was controlling
both the longitudinal and lateral movement of the vehicle.

This subset of the full naturalistic driving dataset served as ground truth for automated lane keeping. In other words,
given the operational characteristics of Autopilot, we know that the vehicle only leaves the lane in two situations: (1)
during automated lane changes and (2) as part of a ``disengagement'' where the driver elects or is forced to take back
control of the vehicle. We have the full enumeration of both scenarios. The latter is of particular interest to the task
of arguing machines, as one indication of a valuable disagreement is one that is associated with a human driver feeling
sufficiently uncomfortable to elect to take back control of the vehicle. There are 6,500 such instances of disengagement
that are used for evaluating the ability of the disagreement function to discover edge cases and challenging driving
scenarios as discussed in \secref{disagreement}.


\begin{figure}[h!]
  \centering
  \includegraphics[width=\columnwidth]{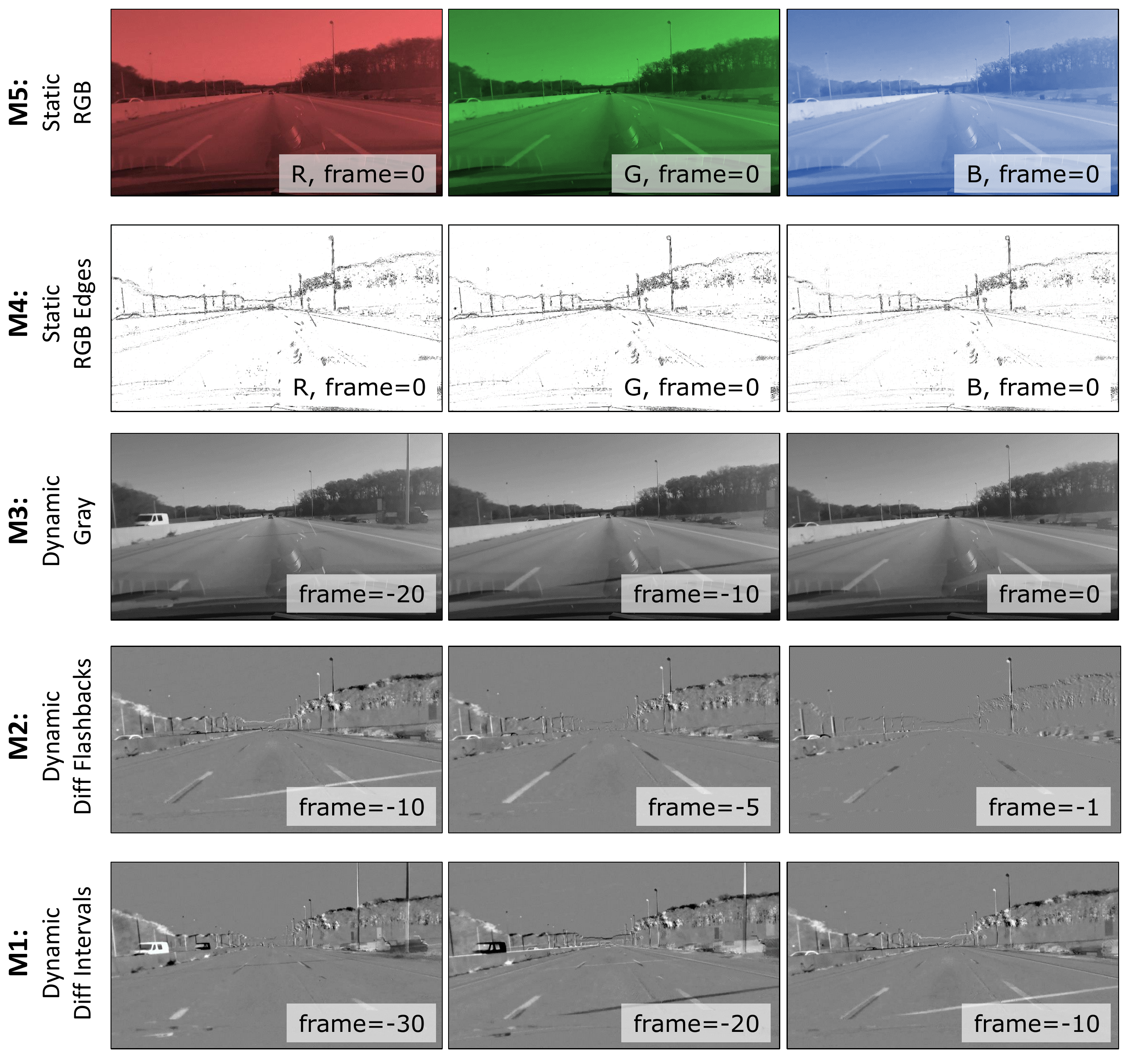}
  \caption{Visualization on one illustrative example of each of the 5 neural network preprocessing models evaluated in
    this paper. See \figref{net-performance} for mean absolute error achieved by each model.}
  \label{fig:net-inputs}
\end{figure}

\begin{figure}[h!]
  \centering
  \includegraphics[width=\columnwidth]{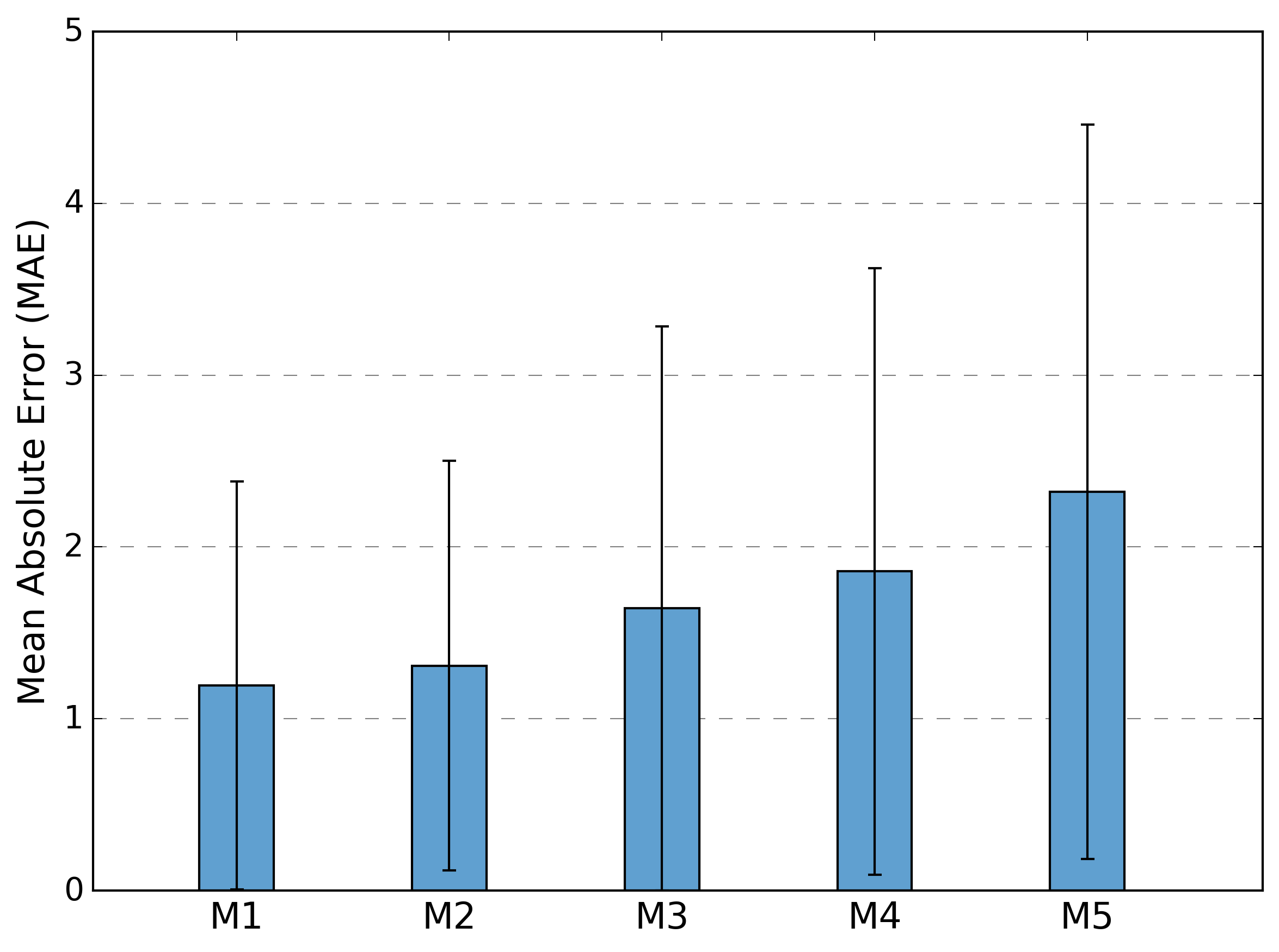}
  \caption{The mean absolute error achieved by each of the 5 models illustrated in \figref{net-inputs}.}
  \label{fig:net-performance}
\end{figure}

\begin{figure}[ht!]
  \centering
  \includegraphics[width=\columnwidth]{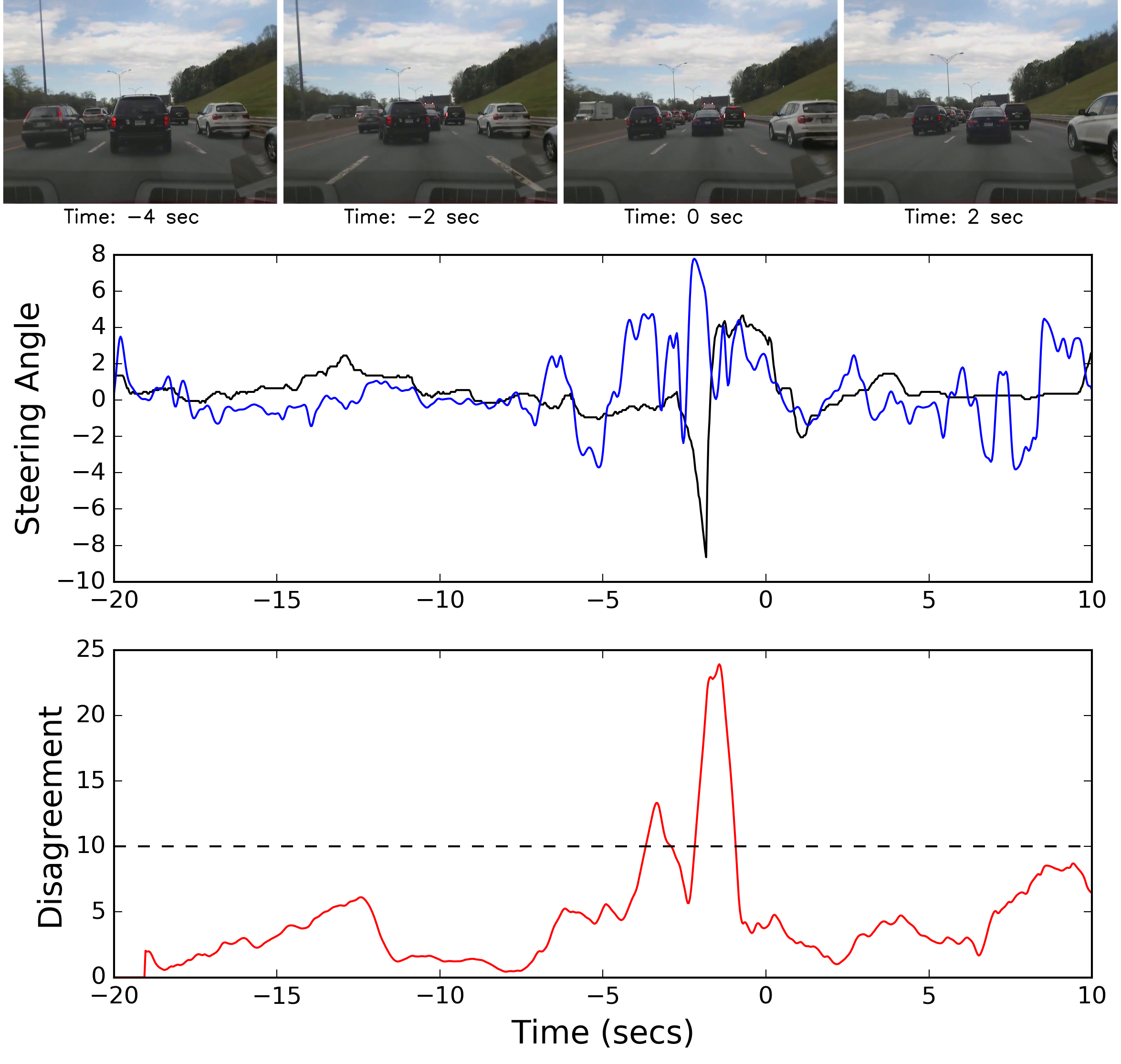}
  \caption{Illustrative example showing snapshots of the forward roadway, plots of the steering angles suggested by the
    primary machine (black line) and secondary machine (blue line), and a plot of the disagreement function along with a
    threshold value of 10 that corresponds to the red circle in \figref{toc-roc}.}
  \label{fig:disagreement-example}
\end{figure}

\subsection{End-to-End Learning of the Steering Task}\label{sec:network}


Our model, which is inspired by \cite{bojarski2016end} uses 5 convolutional layers, the first 3 with a stride of
$2\times2$ and $5\times5$ kernels and the remaining 2 keeping the same stride, while switching to smaller $3\times3$
kernels.  On top of that, we add 4 fully connected layers going down to output sizes of 100, 50, 10, and 1,
respectively. Throughout the net ReLU activations \cite{nair2010rectified} are used on the layers. In addition, we use
Dropout \cite{krizhevsky2012imagenet} as regularization technique on the fully connected layers.
The net is trained using an RMSprop \cite{hinton2012neural} optimizer minimizing the mean squared error between
predicted and actual steering angle.

Since a large part of driving - and therefore also our dataset - consists of going straight, we had to specifically
select input images to remove that imbalance, and resulting bias towards lower steering angle values the net would learn
otherwise. To accomplish this dataset balancing task, we calculate a threshold using the minimum number of available
frames in steering angle ranges of one degree. This threshold is then used within the range of interest of
$[-10^\circ, 10^\circ]$ steering angle to allow at max \textit{threshold} frames get selected to achieve a balance. This
results in about 100,000 training and 50,000 validation frames.

For the input to the neural network we considered 5 different preprocessing methods (see \figref{net-inputs}) -
referenced as M1 - M5 in the following sections - each producing a $256\times144$ image with 3 channels. M5 uses the
method proposed in \cite{bojarski2016end} as a comparison, consisting of the RGB channels of a single frame.  M4 uses
the same single frame, but precomputes edges on each color channel.

To improve the accuracy beyond that, for input methods M1 to M3 we use a temporal component, meaning multiple frames, to
improve situation awareness. M3, in addition to the current frame, also looks 10 and 20 frames back and
provides the grayscale version of them as the input image channels. M2 goes beyond that and, in addition to using
multiple frames as input, also subtracts them from each other, which helps with an implicit input normalization, as well
as automatically highlighting the important moving parts like lane markings. The exact mathematical formulation of the
input is:

\begin{equation}\label{eq:flashback}
  I_t = \{ (F_{t} - F_{t-10}), (F_{t} - F_{t-5}), (F_{t} - F_{t-1})\}
\end{equation}

\vspace{0.1in} \noindent where $I_t$ and $F_t$ are the input to the neural network and the video frame at time $t$. The unit of time is
1 video frame or $33.3$ milliseconds given the 30 fps video used in this work. In \eqref{flashback}, each channel is
based on the current frame, but also incorporates a ``flashback'' to another frame further back.
 
M1 does not use ``flashbacks'', but instead looks at the changes that happened over a series of time segments - each 10
frames long, as follows:

\begin{equation}
  I_t = \{ (F_{i-20}- F_{i-30}), (F_{t-10} - F_{t-20}), (F_{t}- F_{t-10})\}
\end{equation}
 
\vspace{0.1in} To evaluate the network using the different preprocessing methods, we compute the mean absolute steering angle error
over the validation set. The results are shown in \figref{net-performance}. Precomputing edges (M4) already leads to
improved performance over just supplying the RGB image (M5), and providing temporal context (M1-M3) does even better,
with ``flashbacks'' (M2) performing better than just providing multiple frames (M3), and comparing time segments (M1)
performing best. For the evaluation of the disagreement function in \secref{disagreement}, we use M1.

\subsection{Disagreement Function and Edge Case Discovery}\label{sec:disagreement}

The goal for the disagreement function is to compare the steering angle suggested by the ``primary machine'' (Autopilot)
and the ``secondary machine'' (neural network) and based on this comparison to make a binary classification of whether
the current situation is a challenging driving situation or not. The disagreement function can take many forms including
modeling the underlying entropy of the disagreement, but the function computed and evaluated in this work purposefully
took on a simple form through the following process:

\begin{enumerate} 
\item Normalize the steering angle for both the primary and secondary machines to be in $[-1, 1]$ normalized by the
  range $[-10, 10]$ and all angles exceeding the range are set to the range limits.
\item Compute the difference between the normalized steering suggestions and sum them over a window of 1 second (or 30
  samples).
\item Make the binary classification decision based on a disagreement threshold $\delta$.
\end{enumerate}

The metrics used for evaluating the performance of the disagreement system are false accept rate (FAR) and false reject
rate (FRR). Where the detection event of interest is the Autopilot disengagement. In other words, an ``accept'' is a
prediction that this moment in time is likely to be associated with a disengagement and can thus be considered an edge
case for the machine learning system. A ``reject'' is a prediction that this moment in time is not likely to be
associated with a disengagement. In order to compute FAR and FRR measure for a given value of $\delta$, we use
classification windows evenly sampled from disengagement periods and non-disengagement periods. A disengagement period
is defined as the 5 seconds leading up to a disengagement and 1 second following it.

The illustrative example in \figref{disagreement-example} shows the temporal dynamics of the two steering suggestions,
the resulting disagreement, and the role of $\delta$ in marking that moment leading up to the disengagement as an edge
case. The ROC curve in \figref{toc-roc} shows, by varying $\delta$, that the optimal mean error rate is 0.096, and is
achieved when $\delta = 10$. This means that given any 1 second period of Autopilot driving in our test dataset, the
difference function can predict whether a disengagement will happen in the next 5 seconds with 90.4\% accuracy. This is
a promising result that motivates further evaluation of the predictive power of the disagreement function both on
a larger dataset of Autopilot driving and in real-world on-road testing.

\subsection{On-Road Deployment}\label{sec:deployment}

As part of exploring and validating the concept of arguing machines we also built a version that runs real time inside a
car. This system consists of a NVIDIA Jetson TX2 to run the model, a 23 inch high resolution screen for the human
interface attached over the center stack of a Tesla Model S with Autopilot version 1, a custom interface to connect to
the vehicle CAN bus to get its current steering angle and a dashboard-mounted Logitech C920 camera capturing the forward
roadway scene at 720p resolution at 30fps.

The system uses OpenCV's camera capture module \cite{bradski2000opencv} to get a live, real-time video stream of the
road scene from the C920 camera. The captured image is stored in a short-term, dynamic, temporally-sorted buffer
structure and uses that buffer structure to assemble the right combination of frames. In this case, we used our best
network model (M1), where frames that are 10, 20 and 30 frames back in time are combined with the current frame to
compute the input for the neural network. The in-car neural network uses the same network layout as described above,
running an optimized PyTorch \cite{paszke2017pytorch} implementation on the Jetson TX2's Tegra Parker SoC with a Pascal
GPU compute chip. The steering angle computed by the neural network and the one captured from Tesla's autopilot system
are then fed into the actual disagreement measurement routine and additionally displayed on the center stack mounted
screen. In addition, in case of a severe disagreement, the system also displays a ``disagreement detected'' warning on
the same screen.

Even though this system is a proof of concept, it achieves a latency from camera input to screen GUI update of less than
200 milliseconds, while performing neural network inference in real time. During an on-road demonstration
during evening rush hour it appears to work reliably and help to warn the driver of oncoming difficult situations in
multiple instances. The video of the demonstration is available online at \website.




\section{Conclusion}

This work proposes a framework for integrating a human supervisor into the decision making process of a black box AI
system that is tasked with making life critical decision. We demonstrate this framework in two applications: (1) an
illustrative example of image classification and (2) on large-scale real-world semi-autonomous driving data.  For the
first application, we apply this framework to image classification achieving a reduction from 8.0\% to 2.8\% top-5 error
on ImageNet. For the second application, we apply this framework to Tesla Autopilot and demonstrate the ability to
predict 90.4\% of system disengagements that were labeled by human annotators as challenging and needing human
supervision. Finally, we implement, deploy, and demonstrate our system in a Tesla Model S vehicle operating in real-world
conditions.



\bibliographystyle{ACM-Reference-Format}
\bibliography{arguing_machines}

\end{document}